# Few-shot learning for medical text: A systematic review

Yao Ge[1], Yuting Guo[1], Yuan-Chi Yang[1], Mohammed Ali Al-Garadi[1], and Abeed Sarker[1,2,*]

[1]Department of Biomedical Informatics, School of Medicine, Emory University, Atlanta, GA

[2]Department of Biomedical Engineering, Georgia Institute of Technology and Emory University, Atlanta, GA

[*]**Corresponding Author:**
101 Woodruff Circle
Suite 4101
Atlanta GA
30322
abeed@dbmi.emory.edu

**Keywords:**
Natural Language Processing (D009323);
Text Mining (D057225);
Machine Learning (D000069550);
Systematic Review (D000078182);
Medical Informatics (D008490)

** A LaTeX formatted version of this article is available at:

https://drive.google.com/file/d/1oBaMzeL-76W5JoC0jDzmTCApbPIU1SJU/view?usp=sharing

# ABSTRACT

**Objective**
Few-shot learning (FSL) methods require small numbers of labeled instances for training. As many medical topics have limited annotated textual data in practical settings, FSL-based natural language processing (NLP) methods hold substantial promise. We aimed to conduct a systematic review to explore the state of FSL methods for medical NLP.

**Materials and Methods**
We searched for articles published between January 2016 and August 2021 using PubMed/Medline, Embase, ACL Anthology, and IEEE Xplore Digital Library. To identify the latest relevant methods, we also searched other sources such as preprint servers (*eg.*, medRxiv) via Google Scholar. We included all articles that involved FSL and any type of medical text. We abstracted articles based on data source(s), aim(s), training set size(s), primary method(s)/approach(es), and evaluation method(s).

**Results**
31 studies met our inclusion criteria—all published after 2018; 22 (71%) since 2020. Concept extraction/named entity recognition was the most frequently addressed task (13/31; 42%), followed by text classification (10/31; 32%). Twenty-one (68%) studies reconstructed existing datasets to create few-shot scenarios synthetically, and MIMIC-III was the most frequently used dataset (7/31; 23%). Common methods included FSL with attention mechanisms (12/31; 39%), prototypical networks (8/31; 26%), and meta-learning (6/31; 19%).

**Discussion**
Despite the potential for FSL in biomedical NLP, progress has been limited compared to domain-independent FSL. This may be due to the paucity of standardized, public datasets, and the relative underperformance of FSL methods on biomedical topics. Creation and release of specialized datasets for biomedical FSL may aid method development by enabling comparative analyses.

# INTRODUCTION

Few-shot learning (FSL), also referred to as low-shot learning, is a class of machine learning methods that attempt to learn to execute tasks using small numbers (*ie.*, few) of labeled training examples.[1–3] In supervised learning (*ie.*, learning from labeled data) settings with limited training instances, the application of traditional machine learning methods typically leads to overfitting (*ie.*, the learner is incapable of generalizing the characteristics of the training data).[4,5] Learning from small numbers of or even single training instances (referred to as one-shot learning) is challenging for machine learning models, although it is conceptually possible since humans are often capable of generalizing learned concepts using only partial information[6] (*eg.*, recognizing numbers or pictures[3]). Thus, true replication of human behavior by artificial intelligence requires the development of models that can learn to generalize from small numbers of training instances—an objective that FSL aims to achieve.

For many NLP tasks, particularly within the medical domain (*eg.*, for rare or novel diseases), the availability of labeled data is limited. Even when large labeled datasets are created for targeted tasks, due to restrictions associated with data privacy and patient security, it can be difficult or impossible to release or share them if they originate from medical sources, such as electronic health records (EHRs). Oftentimes, there is just not enough data to annotate, and even when data is available, manually annotating them can be time-consuming, error-prone and/or costly. This is particularly true for medical free text as manual annotation requires the annotators to read and interpret texts prior to assigning labels, and often reliable annotations can only be obtained from high-skilled annotators (*eg.*, doctors) and/or multiple rounds of annotations on the same data. Over recent years, deep neural network based approaches (*aka.*, *deep learning*) have seen high adoption, and they have achieved state-of-the-art results in many supervised learning tasks, sometimes achieving human-level performances.[7] However, such methods require large amounts of labeled training data, which restricts their utility to only tasks that have them. Dictionary- or lexicon-based approaches are commonly used in biomedical named entity recognition (NER) or concept detection, particularly in the absence of large, manually annotated datasets. Such approaches utilize lists of biomedical terms (*ie.*, lexicons or dictionaries) to identify relevant expressions in texts, usually via string matching techniques. Lexicon-based methods typically work well compared to deep learning methods for NER tasks such as detecting drugs, doses, and symptoms, when the number of relevant expressions is small and the texts do not contain too many lexical variants (*eg.*, misspellings). However, these approaches are not very scalable and particularly underperform when the number of concepts is large, concepts are expressed using a variety of expressions, and/or concept expressions are ambiguous. This is particularly true for data from sources such as electronic health records (EHRs) and social media, where large numbers of abbreviations and variants exist. The limitations of lexicon-based and deep learning approaches provide motivation for the development of methods, such as FSL, that can effectively learn from small samples of labeled data.

**Few-shot learning, natural language processing and the medical domain**

Early FSL research primarily focused on the field of computer vision, particularly with the goal of replicating how children learn to distinguish objects with minimal or no supervision.[8–10] FSL research progress in NLP has been notably slower, primarily due to greater difficulties posed by natural language data and the lack of unified benchmarks in few-shot NLP.[11] Attaining high machine learning performances has also been challenging in few-shot settings. Unlike images, text-based data often contain ambiguities and connotations that make generalization complicated. The presence of domain-specific terminologies, expressions, and associations in medical texts further exacerbates the difficulties of FSL.[12] Due to the potential utility of FSL in medical NLP, research on the topic is receiving growing attention, and progress has primarily occurred by building on a small set of related but distinct promising categories of approaches.

As only small numbers of labeled examples are available in the training data, prior knowledge, which is the knowledge the learner has before training, plays an indispensable role in FSL.[13] Using prior knowledge, FSL can potentially generalize to new tasks in an effective manner as the small number of training instances are sufficient for fine-tuning the models for the task.[14] Wang et al.[14] divides FSL methods into three categories based on how prior knowledge is used: (i) data, which use prior knowledge to augment training data; (ii) model, which use prior knowledge to constrain hypothesis space; and (iii) algorithm, which use prior knowledge to guide how parameters are obtained.

One set of promising FSL approaches involves *meta-learning* (*aka.*, "*learning to learn*"[15]). Meta-learning has perhaps been the most common framework for FSL, and is a branch of *metacognition*, which is concerned with learning about one's own learning and learning processes.[16] In the classical machine learning framework, training data is used to optimize a model for a specific task, and a separate set is used to evaluate the performance of the trained model. In the meta-learning framework, a model is trained using a set of training *tasks*, not data, and model performance is evaluated on a set of test tasks. In the experimental setting, the learner obtains prior knowledge by incorporating generic knowledge across different tasks (*ie.*, algorithm level prior knowledge). The small number of labeled instances for the target task are then used to fine-tune the model. Figure 1(a) illustrates the meta-learning framework using a simple example—an entity recognition model is trained using different tasks involving news and music data, and is evaluated on a medical task.

Several additional classes of FSL methods have evolved over the years, some building on meta-learning. Ravi and Larochelle,[8] presented a long-short term memory (LSTM) based meta-learner that is trained and customized separately for mini-batches of training data (referred to as *episodes*), rather than as a single model over all the mini-batches. Separately, *matching networks* were recently proposed, and they attempt to use two embedding functions (*ie.*, functions that project data into vector space while capturing relevant semantics)—one for the training sets and one for the test sets—to imitate how humans generalize the knowledge learned from examples. The framework attempts to optimize the two embedding functions from the training (*support sets*) and the validation examples (*query sets*), and attempts to measure how well the trained model can generalize.[9,17] Figure 1(b) illustrates the functionality of matching networks in a simplified manner. A variant of matching networks utilizes active learning by adding a sample selection step that augments the training data by labeling the most beneficial unlabeled sample (*ie.*, model level prior knowledge).

Another related class of FSL approaches known as *metric learning* employs distance-based metrics (*eg.*, nearest neighbor). Given a support set, metric learning methods typically produce weighted nearest neighbor classifiers via non-linear transformations in an embedding space, and the examples in the support set close to the query example (based on the metric applied) are used to make classification decisions, imitating how humans use similar examples or analogies to learn. *Prototypical networks*,[2] yet another similar class of approaches, particularly attempt to address the issue of overfitting due to small training samples by generating prototype representations of classes from the training samples, similar to how humans summarize knowledge learned from examples. Prediction of unknown data samples can be performed by computing distances to the class prototypes (*eg.*, support set means), and choosing the nearest one as the predicted label. Figure 1(c) visually illustrates the functionality of a prototypical network. A semi-supervised variant of prototypical networks applies *soft assignment* on unlabeled samples, and incorporates these as prior knowledge (*ie.*, data level prior knowledge. Transfer learning, a commonly used approach in FSL, also incorporates prior knowledge at the data level as knowledge learned from data in prior tasks are *transferred* to new few-shot tasks.[18]

The problems that these and other FSL methods attempt to solve are closely aligned with the practical challenges faced by many medical NLP tasks. While a number of FSL strategies have been explored for medical texts by distinct research communities (*eg.*, health informatics, computational linguistics), there is currently no review that compares the performances of these strategies or summarizes the current state of the art. There is also no study that has compiled the reported performances of FSL methods on distinct medical NLP data/tasks. We attempt to address these gaps in this systematic review. Specifically, we review FSL methods for medical NLP tasks, and characterize each reviewed article in terms of type of task (*eg.*, text classification, NER), primary aim(s), dataset(s), evaluation metrics, and other relevant aspects. We summarize our findings about FSL methods for medical NLP, and discuss challenges, limitations, opportunities and necessary future efforts for progressing research on the topic.

## MATERIALS AND METHODS
### Search strategy
We followed the Preferred Reporting Items for Systematic Reviews and Meta-Analysis (PRISMA) protocol to conduct this review.[20] FSL for NLP is a relatively recent research topic, so we concentrated on a short time range for our literature search—January 2016 to August 2021. We searched the following bibliographic databases to identify relevant papers: (1) PubMed/Medline, (2) Embase, (3) IEEE Xplore Digital Library, (4) ACL Anthology, and (5) Google Scholar, the latter being a meta-search engine, not a database. We included ACL Anthology (the primary source for the latest NLP research) and IEEE Xplore, in addition to EMBASE and PubMed/Medline, because much of the methodological progress in FSL has been published in non-medical journals and conference proceedings. At the time of searching (September 2021), ACL Anthology hosted 71,290, and IEEE Xplore hosted over 5.4 million articles, although most articles in the latter did not focus on NLP or medicine. Over recent years, preprint servers have emerged as major sources of the latest information regarding research progress in computer science and NLP, and we used Google Scholar primarily as a medium for searching these preprint servers or published papers from other sources. Note that we also searched the ACM Digital Library[*], but discovered no additional article. Hence, we do not report it as a data source for our review.

We applied marginally different search strategies depending on the database to account for the differences in their contents. We used three types of queries:

1. Queries focusing on the technical field of research (phrases included: 'natural language processing', 'text mining', 'text classification', 'named entity recognition', and 'concept extraction');
2. Queries focusing on the learning strategy (phrases included: 'few-shot', 'low-shot', 'one-shot', and 'zero-shot'); and
3. Queries focusing on the domain of interest (phrases included: 'medical', 'clinical', 'biomedical', 'health', 'health-related').

All articles on PubMed and Embase fall within the broader biomedical domain, so we used combinations of the phrases in 1 and 2 above for searching these two databases, leaving out the phrases in 3. All articles in the ACL Anthology involve NLP, so we used phrases from 2 and 3 for this source. For IEEE Xplore and Google Scholar, the articles can be from any domain and on any topic, so we used combinations of all three sets of phrases for searching. PubMed, Embase, and IEEE only returned articles that entirely matched the queries. However, ACL Anthology and Google Scholar retrieved larger sets of articles and ranked them by relevance. For ACL Anthology, the articles retrieved were reviewed sequentially in decreasing order of relevance. For each query combination, we continued reviewing candidate articles until we came across at least two pages (about 20 articles) of no relevant articles, at which point we decided that no relevant articles would be found in the following pages. Since FSL is a relatively new research area, we anticipated that there would be some relevant research papers that are not yet indexed in PubMed, Embase, IEEE Xplore or ACL Anthology. Specifically, preprint servers such as arXiv, bioRxiv and medRxiv are very popular among machine learning and NLP researchers as they enable the publication of the latest research progress early. We used Google Scholar as an auxiliary search engine to identify potentially relevant articles indexed in such preprint servers or other sources (*eg.*, Open Review[†]). Google Scholar, like ACL Anthology, sorts returned articles by relevance, but the total number of articles returned is much larger. For this search engine, therefore, we reviewed the top 40 articles returned by each query combination, excluding those that were retrieved from the other databases.

**Study selection and exclusion criteria**
All articles shortlisted from initial searches were screened for eligibility by two authors of the manuscript (YGe and AS). We removed duplicate articles and those that either did not include at least one dataset from the biomedical domain, or did not involve NLP. While it was always possible to identify the technical field/topic (NLP or not) from the titles and abstracts, to determine domain, we had to review full articles because a subset of papers included multiple datasets, and only some of these datasets were from the medical domain. We excluded papers if none of the datasets were related to medicine/health, or did not explicitly focus on few/low-shot settings, and reviewed the remaining articles.

**Data abstraction and synthesis**
We abstracted the following details from each article, if available: publication year, data source, primary research aim(s), training set size(s), number of entities/classes, entity type for training, entity type for evaluation/testing, primary method(s), and evaluation methodology. For studies including data from multiple sources, we only abstracted those related to health/medicine. In terms of primary aim(s), some studies reported multiple objectives, and we abstracted all the NLP-oriented ones (*eg.*, text classification, concept extraction). With respect to training set sizes, we abstracted information about the number of instances that were used for training, and, if applicable, how larger datasets were *reconstructed* to create few-shot samples. We also extracted the number of labels for each study/task; for NER/concept extraction methods, we identified the number of entities/concepts, and for classification, we identified the type of classification (*ie.*, multi-label or multi-class) along with the number of classes.

[*]https://dl.acm.org/. Last accessed March 9, 2022.
[†]https://openreview.net/. Last accessed March 9, 2022.

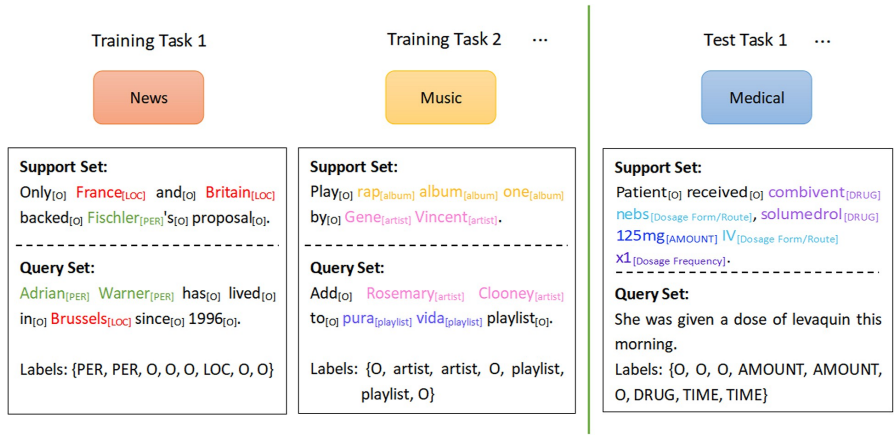

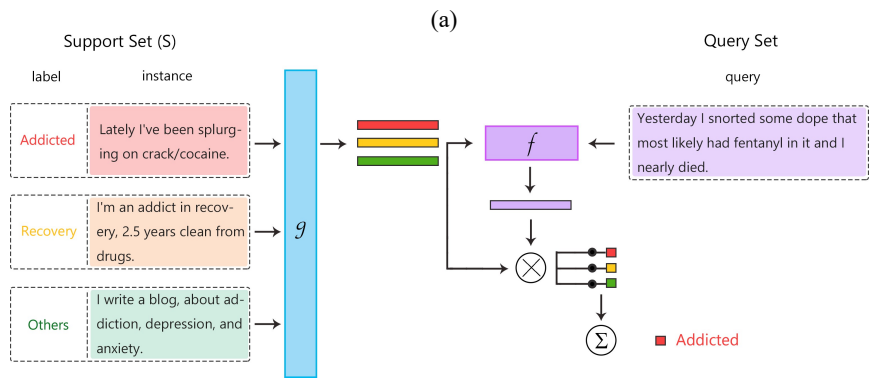

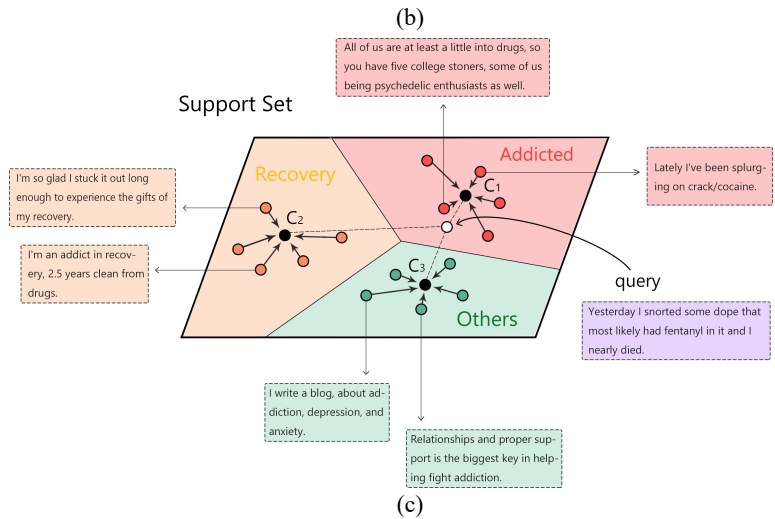

**Figure 1.** Architectures of three popular few-shot learning methodologies. (a) Meta-learning: each task mimics the few-shot scenario, and can be completely non-overlapping. Support sets are used to train; query sets are used to evaluate the model. (b) Matching networks: a small support set contains some instances with their labels (one instance per label in the figure). Given a query, the goal is to calculate a value that indicates if the instance is an example of a given class. For a similarity metric, two embedding functions f() and g() need to take similarity on feature space. The function f() is a neural network and then the embedding function g() applied to each instance to process the kernel for each support set. (c) Prototypical network: a class's prototype is the mean of its support set in the embedding space. Given a query, its distance to each class's prototype is computed to decide its label. Note: (b) and (c) use the DASH 2020 Drug Data[19])

We also noted down the training domain(s) and test/evaluation domain(s) for each few-shot method, when applicable. Abstracting primary approach(es) and evaluation methodology was more challenging due to the complexities of some of the model implementations, and we reviewed and summarized the descriptions provided in each paper. For evaluation, we abstracted evaluation strategies and reported performances.

## RESULTS

### Data collection results

31 studies met our inclusion criteria. Initial searches retrieved 1241 articles from PubMed, Embase, IEEE Xplore and ACL Anthology, and an additional 459 from Google Scholar. Figure 2 presents the screening procedures and numbers at each stage. After initial filtering, we reviewed 46 full-text articles for eligibility, and excluded 15 from the final review. The first included study was from 2018, and most articles (22/31; 71%) were from 2020 and 2021, although for the latter year, only studies published prior to August 31 were included.

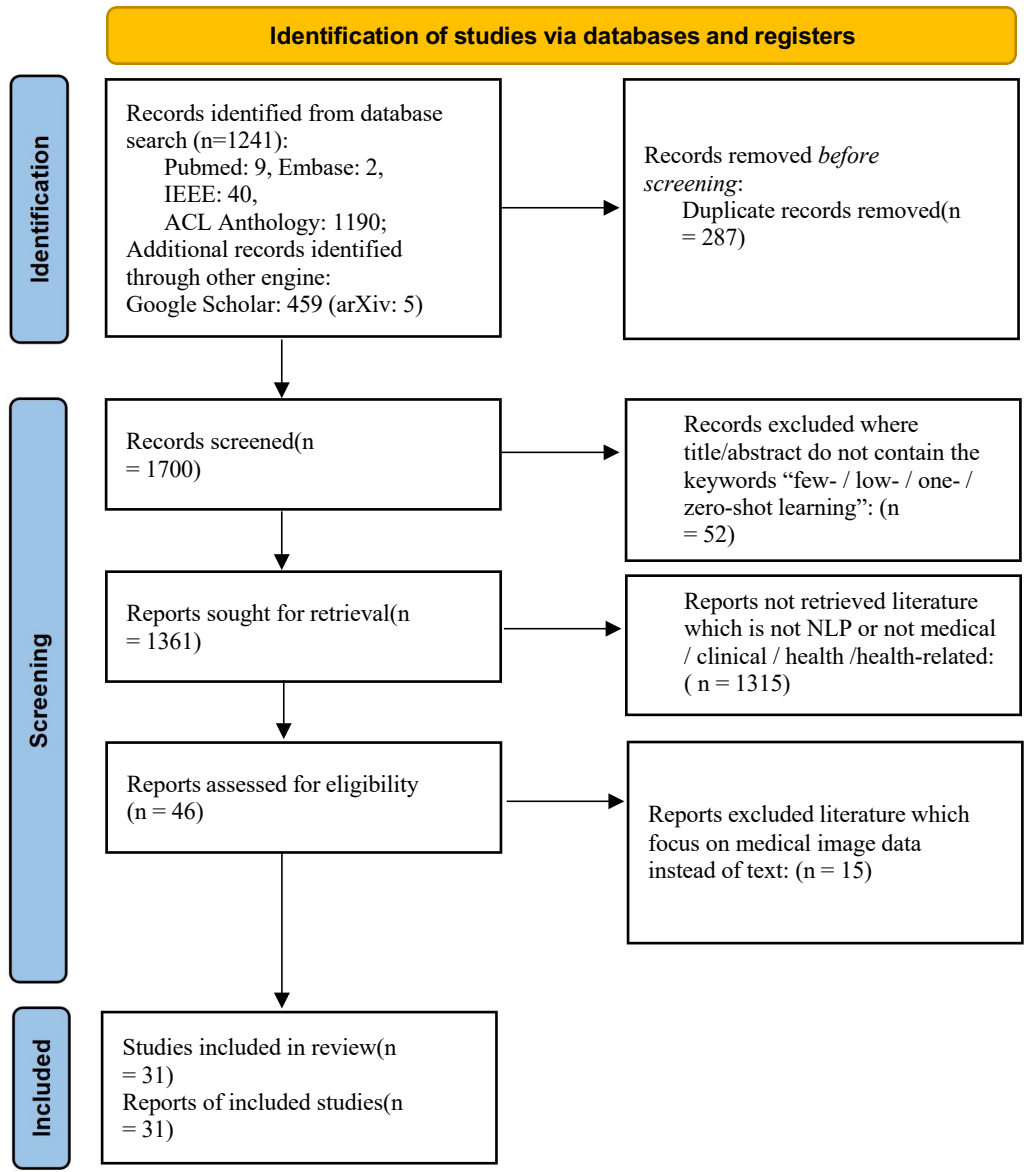

**Figure 2.** PRISMA flow diagram for the process of depicting the number of articles at each stage of collection and filtering.

| Study | Year | Data source | Research aim | Size of training set | Number of entities / classes | Entity type of training domain | Entity type of test domain |
|---|---|---|---|---|---|---|---|
| Anthony Rios and Ramakanth Kavuluru[21] | 2018 | MIMIC II[22] and MIMIC III[23] | Multi-label Text Classification | Frequent group (all labels that occur >5 times), the few-shot group (labels that occur between 1 and 5 times), and the zero-shot group (labels that never occur in the training dataset), reconstructed | Not mentioned * | Medical, discharge summaries annotated with a set of ICD-9 diagnosis and procedure labels | Medical, discharge summaries annotated with a set of ICD-9 diagnosis and procedure labels |
| Anthony Rios and Ramakanth Kavuluru[24] | 2018 | MIMIC II[22] and MIMIC III[23] | Multi-label Text Classification | Original dataset, with no reconstruction | Not mentioned * | Medical | Medical |
| Hofer et al.[12] | 2018 | i2b2 2009[25], 2010[26], 2012[27]; CoNLL-2003[28]; BioNLP-2016[29]; MIMIC-III[23]; UK CRIS[30,31] | NER | 10-shot | Not mentioned * | Medical and nonmedical (e.g, news) | Medical |
| Pham et al.[32] | 2018 | The Europarl datasets[33], and IWSLT17[34] for English→Spanish; the UFAL Medical Corpus and HIML2017 dataset ‡ for German→English | Neural Machine Translation (NMT) | One-Shot | N/A † | German→English: medical; English→Spanish, the proceedings of the European Parliament and data from TED | German→English: medical; English→Spanish: data from TED |
| Yan et al.[35] | 2018 | Multigames dataset[36], HCR dataset[37] and SS-Tweet dataset[38], SemEval-2013 Dataset (SemEval b)[39] | Text Classification | Few-shot, but reconstructed | Multigames : 3 Hcr: 5 SS-Tweet: 3 SemEval-2013 Dataset: 3 | Tweets about sentiment and games, and health Care Reform (HCR) data | Tweets about sentiment and games, and Health Care Reform (HCR) data |
| Manousogiannis et al.[40] | 2019 | Tweets (provided by SMM4H 2019)[41] | NER | Original dataset, with no reconstruction | 1, ADR with 319 MEDDRA codes | Medical (ADR) | Medical (ADR) |
| Gao et al.[42] | 2019 | Based on FewRel dataset[43], this paper propose FewRel 2.0 by constructing a new test set from the biomedical domain | Relation Classification | 5-Way 1-Shot / 5-Way 5-Shot / 10-Way 1-Shot / 10-Way 5-Shot | 25 | Wikipedia corpus and Wikidata knowledge bases | Biomedical literature with UMLS, a large-scale knowledge base in the biomedical sciences |

| Study | Year | Data source | Research aim | Size of training set | Number of entities / classes | Entity type of training domain | Entity type of test domain |
|---|---|---|---|---|---|---|---|
| Alicia Lara-Clares and Ana Garcia-Serrano[44] | 2019 | MEDDOCAN shared task dataset[45] | NER | 500 clinical cases, with no reconstruction | 29 | Clinical | Clinical |
| Ferré et al.[46] | 2019 | BB-norm dataset from the Bacteria Biotope 2019 Task[47] | Entity Normalization | Original dataset with no reconstruction and zero-shot | Not mentioned * | Biological | Biological |
| Hou et al.[48] | 2020 | Snips dataset[49] | Slot Tagging (NER) | 1-shot and 5-shot | 7 | Six of Weather, Music, PlayList, Book (including biomedical), Search Screen (including biomedical), Restaurant and Creative Work. | The remaining one |
| Sharaf et al.[50] | 2020 | ten different datasets collected from the Open Parallel Corpus (OPUS)[51] | Neural Machine Translation (NMT) | Sizes ranging from 4k to 64k training words (200 to 3200 sentences), but reconstructed | N/A † | Bible, European Central Bank, KDE, Quran, WMT news test sets, Books, European Medicines Agency (EMEA), Global Voices, Medical (ufal-Med), TED talks | Bible, European Central Bank, KDE, Quran, WMT news test sets, Books, European Medicines Agency (EMEA), Global Voices, Medical (ufal-Med), TED talks |
| Lu et al.[52] | 2020 | MIMIC II[22] and MIMIC III[23], and EU legislation dataset[53] | Multi-label Text Classification | 5-shot for MIMIC II and III, 50-shot for EU legislation | MIMIC II: 9 MIMIC III: 15 EU legislation: 5 | Medical | Medical |

| Study | Year | Data source | Research aim | Size of training set | Number of entities / classes | Entity type of training domain | Entity type of test domain |
|---|---|---|---|---|---|---|---|
| Chen Jia and Yue Zhang[54] | 2020 | BioNLP13PC and BioNLP-13CG[55], CoNLL-2003 English dataset[28], Broad Twitter dataset[56], Twitter dataset[57] and CBS SciTech News dataset[58] | NER | Four few-shot (reconstructed) and zero-shot | CoNLL: 4 Broad Twitter: 3 Twitter: 4 BioNLP13PC: >=3 BioNLP13CG: >=3 CBS News: 4 | For the BioNLP dataset, BioNLP13PC as the source domain dataset; In the Broad Twitter dataset, the CoNLL-2003 as the source domain dataset; In the Twitter dataset, the CoNLL-2003 as the source domain dataset | for the BioNLP dataset, BioNLP13CG is used as the target domain dataset; In the Broad Twitter dataset, Broad Twitter is used as the target domain dataset; In the Twitter dataset, Twitter is used as the target domain dataset |
| Chalkidis et al.[59] | 2020 | EURLEX57K[53], MIMIC III[23] and AMAZON13K[60] | Multi-label Text Classification | The labels are divided into frequent (>50), few-shot (≤0), and zero-shot | Not mentioned * | English legislative documents, English discharge summaries from US hospitals, English product descriptions from Amazon | English legislative documents, English discharge summaries from US hospitals, English product descriptions from Amazon |
| Brandon Lwowski and Paul Rad[61] | 2020 | Tweets about COVID-19[62] | Text Classification | 100 tweets with no reconstruction | 4 | Tweets about COVID-19 | Tweets about COVID-19 |
| Hou et al.[11] | 2020 | dialogue utterances from the AIUI open dialogue platform of iFlytek[§] | Dialogue Language Understanding: includes two sub-tasks: Intent Detection (classification) and Slot Tagging (sequence labeling) | 1-shot, 3-shot, 5-shot and 10-shot | Train Domains: 45 Dev Domains: 5 Test Domains: 9 | General dialogue (including health domain) | General dialogue (including virus-Search domain) |
| Chen et al.[63] | 2020 | WIKIBIO dataset[64] | Natural Language Generation (NLG) | Dataset sizes: 50, 100, 200 and 500, with no reconstruction | N/A [†] | Books, Songs and Human domain (including biomedical) | Books, Songs and Human domain (including biomedical) |

| Study | Year | Data source | Research aim | Size of training set | Number of entities / classes | Entity type of training domain | Entity type of test domain |
|---|---|---|---|---|---|---|---|
| Vaci et al.[65] | 2020 | UK-CRIS system that provides a means of searching and analysing deidentified clinical case records from 12 National Health Service Mental Health Trusts[30, 31] | NER | Original dataset, with no reconstruction | 7 | Clinical | Clinical |
| Huang et al.[66] | 2020 | 10 public datasets | NER | 5-shot, 10%, 100% | CoNLL: 4<br>Onto: 18<br>WikiGold: 4<br>WNUT: 6<br>Movie: 12<br>Restaurant: 8<br>SNIPS: 53<br>ATIS: 79<br>Multiwoz: 14<br>I2B2: 23 | 10 public datasets, different domains | 10 public datasets, different domains |
| Chen et al.[67] | 2020 | MRI image dataset and MRI text reports¶ | Text Classificaition | Original dataset, with no reconstruction | Not mentioned * | MRI data | MRI data |
| Yin et al.[68] | 2020 | MLEE[69] and BioNLP'13-GE[55] | Sequence Tagging (NER) | 5-way-10-shot, 5-way-15-shot and 5-way-20-shot | 5 | Biological event | Biological event |
| Goodwin et al.[70] | 2020 | TensorFlow DataSets catalogue‖ | Abstractive Summarization | Zero-shot and 10-shot | N/A † | 3 general domain & 1 consumer health | 3 general domain & 1 consumer health |
| Yi Yang and Arzoo Katiyar[71] | 2020 | OntoNotes 5.0[72], CoNLL-2003[28], I2B2 2014[73], WNUT 2017[74] | NER | 1-shot and 5-shot | Onto: 18<br>CoNLL: 4<br>I2B2-14: 23<br>WNUT: 6 | Three of general, news, medical and social media | The remaining one of general, news, medical and social media |
| Mareike Hartmann and Anders Søgaard[75] | 2021 | The IULA dataset (Spanish)[76], The NUBES dataset (Spanish)[77], The FRENCH dataset[78]; all from the hospitals. And other Negation Scope Resolution datasets | NER | Zero-shot, with no reconstruction | 1, Negation | No training data for the clinical datasets | Clinical |
| Pieter Fivez and Simon Suster[79] | 2021 | SNOMED-CT1** disorder names as biomedical synonym sets and ICD-10 | Name Normalization | Zero-shot, with no reconstruction | N/A † | Biomedical | Biomedical |

| Study | Year | Data source | Research aim | Size of training set | Number of entities / classes | Entity type of training domain | Entity type of test domain |
|---|---|---|---|---|---|---|---|
| Lu et al.[80] | 2021 | Constructed and shared a novel dataset [††] based on Weibo for the research of few-shot rumor detection, and use PHEME dataset[81] | Rumor Detection (NER) | For the Weibo dataset: 2-way 3-event 5-shot 9-query; for PHEME dataset: 2-way 2-event 5-shot 9-query | Weibo: 14 PHEME: 5 | Source posts and comments from Sina Weibo related to COVID-19 | Source posts and comments from Sina Weibo related to COVID-19 |
| Ma et al.[82] | 2021 | CCLE, CERES-correctedCRISPR gene disruption scores, GDSC1000 dataset, PDTC dataset and PDX dataset[‡‡] | Drug-response Predictions | 1-shot, 2-shot, 5-shot and 10-shot | N/A[†] | Biomedical | Biomedical |
| Kormilitzin et al.[83] | 2021 | MIMIC-III[23] and UK-CRIS datasets[30,31] | NER | 25%, 50%, 75% and 100% of the training set, with no reconstruction | 7 | Electronic health record | Electronic health record |
| Guo et al.[84] | 2021 | Abstracts of biomedicalliteratures (from relation extraction task of BioNLP Shared Task 2011 and 2019[47]) and structured biological datasets | NER | 100%, 75%, 50%, 25%, 0% of training set, with no reconstruction | Not mentioned [*] | Biomedical entities | Biomedical entities |
| Lee et al.[85] | 2021 | COVID19-Scientific[86], COVID19-Social[87] (fact-checked by journalists from a website called Politi-fact.com), FEVER[88] (Fact Extraction and Verification, generated by altering sentences extracted from Wikipedia to promote research onfact-checking systems) | Fact-Checking (close to Text Classification) | 2-shot, 10-shot and 50-shot | Not mentioned [*] | Facts about COVID-19 | Facts about COVID-19 |

| Study | Year | Data source | Research aim | Size of training set | Number of entities / classes | Entity type of training domain | Entity type of test domain |
|---|---|---|---|---|---|---|---|
| Pieter Fivez and Simon Suster[89] | 2021 | extract sets of high-level concepts and their constituent names from 2 large-scale hierarchies of disorder concepts, ICD-10 and SNOMED-CT** | Name Normalization | 15-shot | †  N/A | Biomedical | Biomedical |

**Table 1.** Articles published for few-shot learning on medical data, their publication years, data sources, search engine, which downstream tasks the literature focus, size of the training set (number of the shot), number of the entities (for few-shot NER tasks), number of the classes (for few-shot classification tasks), type of data in training domain, type of data in test domain.

* The research aim of this paper is text classification or NER, but the size of training set is not mentioned in the paper.

† The research aim of this paper is neither text classification nor NER.

‡ UFAL Medical Corpus v.1.0 and HIML2017 dataset: http://aiui.xfyun.cn/index-aiui. Last accessed March 9, 2022.

§ iFlytek: http://aiui.xfyun.cn/index-aiui. Last accessed March 9, 2022.

¶ Those datasets are not released.

∥ TensorFlow DataSets: https://www.tensorflow.org/datasets. Last accessed March 9, 2022.

** SNOMED-CT1: https://www.snomed.org. Last accessed March 9, 2022.

†† A novel dataset proposed by this paper: https://github.com/jncsnlp/Sina-Weibo-Rumors-for-few-shot-learning-research. Last accessed March 9, 2022.

‡‡ Links are provided in the original paper.

**Dimensions of characterization**

Table 1 summarizes some fundamental characteristics of each study (authors, year, data source, retrieval search engine, number of entities/classes, research aims, training set sizes, and training and evaluation entity types). In terms of training data sizes, 7/31 (23%) studies included zero-shot scenarios (*ie.*, prediction without any labeled instances) into their research scope, including two on zero-shot learning only. 1-shot, 5-shot, and 10-shot were the most common '*shot*' settings, representing 12/31 (39%) of the reviewed studies. 6/31 (19%) reviewed studies used samples of larger datasets for training, often specified in percentages (*eg.*, 25%, 50%). 3/31 (10%) studies did not explicitly specify shot values. 2 studies did not perform experiments in accordance with traditional few-shot scenarios, and divided all labels into three categories according to the frequency of occurrences (frequent group contained all labels occurring more than 5 times; few-shot group contained labels occurring between 1 and 5 times; and the zero-shot group included labels that never occurred in the training dataset), causing some labels to have large numbers of annotated samples. 11/13 (85%) few-shot NER tasks explicitly mentioned the number of entity types. For few-shot classification, 50% (5/10) specified the approximate number of classes. 7/31 (23%) studies involved cross-domain transfer, with different domains of training and test/evaluation data. In most cases, however, the training sets and test sets used were from the same domain.

| Study | Research aim | Primary approach(es) | Evaluation methodology |
|---|---|---|---|
| Anthony Rios and Ramakanth Kavuluru[21] | Multi-label Text Classification | Propose and evaluate a neural architecture suitable for handling few- and zero-shot labels in the multi-label setting where the output label space satisfies two constraints: (1). the labels are connected forming a DAG and (2). each label has a brief natural language descriptor. | R@5 and R@10 (Recall), P@10 (Precision), Macro-$F_1$ scores |
| Anthony Rios and Ramakanth Kavuluru[24] | Multi-label Text Classification | Propose a novel semi-parametric neural matching network for diagnosis/procedure code prediction from EMR narratives. | Precision, Recall, $F_1$-scores, AUC (PR), AUC (ROC), P@k, R@k |
| Hofer et al.[12] | NER | Five improvements on NER tasks when only 10 annotated examples are available: 1.Layer-wise initialization with pre-trained weights (single pre-training); 2.Hyperparameter tuning; 3.Combining pre-training data; 4.Custom word embeddings; 5. Optimizing out-of-vocabulary (OOV) words. | $F_1$-score |
| Pham et al. [32] | Neural Machine Translation (NMT) | Present a generic approach to use phrase-based models to simulate Experts to complement neural machine translation models show that the model can be trained to copy the annotations into the output consistently. | BLEU score, SUGGESTION SUG) and SUGGESTION ACCURACY (SAC) ( |

| Study | Research aim | Primary approach(es) | Evaluation methodology |
|---|---|---|---|
| Yan et al.[35] | Text Classification | Propose a short text classification framework based on Siamese CNNs and few-shot learning, which will learn the discriminative text encoding so as to help classifiers distinguish those obscure or informal sentence. The different sentence structures and different descriptions of a topic will be learned by few-shot learning strategy to improve the classifier's generalization. | Accuracy |
| Manousogiannis et al.[40] | Concept Extraction | Propose a simple Few-Shot learning approach, based on pre-trained word embeddings and data from the UMLS, combined with the provided training data. | Relaxed and strict Precision/Recall /$F_1$-scores |
| Gao et al.[42] | Relation Classification | Propose FewRel 2.0, a new task containing two real-world issues that FewRel ignores: (1) few-shot domain adaptation, and (2) few-shot none-of-the-above detection. This work is based in the Few-shot Learning | Accuracy |
| Alicia Lara-Clares and Ana Garcia-Serrano[44] | NER | Model to learn high level features. propose a hybrid Bi-LSTM CNN model adding a Part-of-Speech (POS) tagging layer, that is, information about multi-word entities. And use wikipedia2vec to automatically extract and classify keywords. | $F_1$-score |
| Ferré et al.[46] | Entity Normalization | Propose C-Norm, a new neural approach which synergistically combines standard and weak supervision, ontological knowledge integration and distributional semantics. | The official evaluation tool of the BB-norm task: a similarity score and a strict exact match score. |
| Hou et al.[48] | Slot tagging (NER) | Proposed Collapsed Dependency Transfer and Label-enhanced Task-adaptive Projection Network 1.A collapsed dependency transfer mechanism into CRF to transfer abstract label dependency patterns as transition scores 2.The emission score of CRF: word's similarity to the representation of each label. 3. A Label-enhanced Task-Adaptive Projection Network (L-TapNet) based on TapNet, by leveraging label name semantics in representing labels. | 1.Cross-validate the models on different domains. One target domain for testing, one domain for development, and rest domains as source domains for training. 2. Evaluate $F_1$-scores within each few-shot episode, and average 100 $F_1$-scores from all 100 episodes as the final result |
| Sharaf et al.[50] | Neural Machine Translation (NMT) | Frame the adaptation of NMT systems as a meta-learning problem, where can learn to adapt to new unseen domains based on simulated offline meta-training domain adaptation tasks. | Use BLEU, measure case-sensitive de-tokenized BLEU with SacreBLEU. |
| Lu et al.[52] | Multi-label Text Classification | Present a simple multi-graph aggregation model that fuses knowledge from multiple label graphs encoding different semantic label relationships in order to study how the aggregated knowledge can benefit multi-label zero/few-shot document classification. | Recall@K and nDCG@K. K was set to 10 for MiMIC-II/III and 5 for URLEX57K. |
| Chen Jia and Yue Zhang[54] | NER | The method creates distinct feature distributions for each entity type across domains, which can give better transfer learning power compared to representation networks that do not explicitly differentiate entity types. | $F_1$-score |

| Study | Research aim | Primary approach(es) | Evaluation methodology |
|---|---|---|---|
| Chalkidis et al.[59] | Multi-label Text Classification | 1. Hierarchical methods based on Probabilistic Label Trees (PLTs); 2. Combines BERT with LWAN; 3. Investigate the use of structural information from the label hierarchy in LWAN. Leverage the label hierarchy to improve few and zero-shot learning. | R-Precision@K (RP@K), a top-K version of R-Precision of each document, and nDCG@K |
| Brandon Lwowski and Paul Rad[61] | Text Classification | Propose a self-supervised learning algorithm to monitor COVID-19 Twitter using an autoencoder to learn the latent representations and then transfer the knowledge to COVID-19 Infection classifier by fine-tuning the Multi-Layer Perceptron (MLP) using few-shot learning. | Accuracy, Precision, Recall and $F_1$-score |
| Hou et al.[11] | Dialogue Language Understanding: includes two sub-tasks: Intent Detection (classification) and Slot Tagging (sequence labeling) | Present FewJoint, a novel Few-Shot Learning benchmark for NLP. This benchmark introduces few-shot joint dialogue language understanding, which additionally covers the structure prediction and multi-task reliance problems. | Intent Accuracy, Slot $F_1$-score, Sentence Accuracy |
| Chen et al.[63] | Natural Language Generation (NLG) | The design of the model architecture is based on two aspects: content selection from input data and language modeling to compose coherent sentences, which can be acquired from prior knowledge. | BLEU-4, ROUGE-4 (F-measure) follow the same trend with BLEU-4 |
| Vaci et al.[65] | Concept Extraction | Used a combination of methods to extract salient information from electronic health records. First, clinical experts define the information of interest and subsequently build the training and testing corpora for statistical models. Second, built and fine-tuned the statistical models using active learning procedures. | Precision, Recall and $F_1$-score |
| Huang et al.[66] | NER | Present the first systematic study for few-shot NER, a problem that is previously little explored in the literature. Three distinctive schemes and their combinations are investigated; perform comprehensive comparisons of these schemes on 10 public NER datasets from different domains; Compared with existing methods on few-shot and training-free NER settings, the proposed schemes achieve SoTA performance despite their simplicity. | $F_1$-score |
| Chen et al.[67] | Classification | Propose a classification and diagnosis method for Alzheimer's patients based on multi-modal feature fusion and small sample learning. And then the compressed interactive network is used to explicitly fuse the extracted features at the vector level. Finally, the KNN attention pooling layer and the convolutional network are used to construct a small sample learning network to classify the patient diagnosis data. | Accuracy and $F_1$-score |

| Study | Research aim | Primary approach(es) | Evaluation methodology |
|---|---|---|---|
| Yin et al.[68] | Sequence Tagging (NER) | Mainly adopt the prototypical network, and use the relation module as the distance measurement function to model the task of biomedical event trigger identification. In addition, in order to make full use of the external knowledge base to learn the complex biological context, we introduced a self-attention mechanism. | $F_1$-score |
| Goodwin et al.[70] | Abstractive Summarization | Compare the summarization quality produced by three state-of-the-art transformer-based models: BART, T5, and PEGASUS. | ROUGE-1, ROUGE-2, and ROUGE-L $F_1$-scores, BLEU-4 and Repetition Rate |
| Yi Yang and Arzoo Katiyar[71] | NER | Propose STRUCTSHOT. 1.Use contextual representation to represent each token, uses a nearest neighbor (NN) classifier and a Viterbi decoder for prediction. 2. Test systems on both identifying new types of entities in the source domain as well as identifying new types of entities in various target domains in one-shot and five-shot settings. | $F_1$-score |
| Mareike Hartmann and Anders Søgaard[75] | Concept Extraction | Present a universal approach to multilingual negation scope resolution, and study an approach for zero-shot cross-lingual transfer for negation scope resolution in clinical text, exploiting data from disparate sources by data concatenation, or in an MTL setup. | Two widely used evaluation metrics for negation scope prediction: Percentage of correct spans (PCS) and $F_1$-score over scope tokens |
| Pieter Fivez and Simon Suster[79] | Name Normalization | Take a next step towards truly robust representations, which capture more domain-specific semantics while remaining universally applicable across different biomedical corpora and domains. Use conceptual grounding constraints which more effectively align encoded names to pretrained embeddings of their concept identifiers. | For synonym retrieval: Mean average precision (mAP) over all synonyms. For concept mapping, Accuracy (Acc) and Mean reciprocal rank (MRR) of the highest ranked correct synonym. |
| Lu et al.[80] | Rumor Detection | Collect and contribute a publicly available rumor dataset that is suitable for few-shot learning from Sina Weibo. And introduce a few-shot learning-based multi-modality fusion model named COMFUSE for COVID-19 rumor detection, including text embeddings modules with pre-trained BERT model, feature extraction module with multilayer Bi-GRUs, multi-modality feature fusion module with a fusion layer, and meta-learning based few-shot learning paradigm for rumor detection. | Accuracy |
| Ma et al.[82] | Drug-response Predictions | Applied the few-shot learning paradigm to three context-transfer challenges: (1) transfer of a predictive model learned in one tissue type to the distinct contexts of other tissues; (2) transfer of a predictive model learned in tumor cell lines to patient-derived tumor cell (PDTC) cultures in vitro; and (3) transfer of a predictive model learned in tumor cell lines to the context of patient-derived tumor xenografts (PDXs) in mice in vivo. | Accuracy, Pearson's correlation, AUC |

| Study | Research aim | Primary approach(es) | Evaluation methodology |
|---|---|---|---|
| Kormilitzin et al.[83] | NER | First, the underlying deep neural network language model was pre-trained in a self-supervised manner using the cloze-style approach. Second, using the weak-supervision method, developed synthetic training data with noisy labels. Lastly, incorporated all ingredients into an active learning approach. | Accuracy, Precision, Recall and $F_1$-score |
| Guo et al.[84] | Extract Entity | Proposes BioGraphSAGE model, a Siamese graph neural network with structured databases as domain knowledge to extract biological entity relations from literatures. | Precision (P-value), Recall (R-value) and $F_1$-score |
| Lee et al.[85] | Relations Fact-Checking (close to Text Classification) | Propose a novel way of leveraging the perplexity score from LMs for the few-shot fact-checking task and demonstrate the effectiveness of the perplexity-based approach in the few-shot setting. | Accuracy and the Macro-$F_1$-score |
| Pieter Fivez and Simon Suster[89] | Name Normalization | Explore a scalable few-shot learning approach for robust biomedical name representations which is orthogonal to this paradigm. And use more general higher-level concepts which span a large range of fine-grained concepts. | Spearman's rank correlation coefficient between human judgments and similarity scores of name embeddings, reported on semantic similarity (sim) and relatedness (rel) benchmark |

**Table 2.** A summary table showing primary few-shot approaches and evaluation methodologies

Table 2 provides summaries of the methods proposed, and the evaluation strategies. Variants of neural network based (deep learning) algorithms, such as Siamese Convolutional Neural Networks (an artificial neural network, which processes two different input vectors simultaneously with the same weights to compute a comparable output vector),[35] were the most common. Only 3/31 (10%) articles proposed new datasets, and 2/31 (7%) presented benchmarks for comparing multiple few-shot methods. Evaluation strategies had considerably less diversity. Almost all evaluation methodologies for classification tasks involved standard metrics such as accuracy, precision, recall, and $F_1$-scores, and NER tasks mainly relied on $F_1$-scores only.

**Data characteristics**

We grouped the datasets used into three categories: (i) publicly downloadable (de-identified) data; (ii) datasets from shared tasks; and (iii) new datasets specifically created for the target tasks. We found that datasets belonging to (ii) and (iii) were particularly difficult to obtain—shared task data are often difficult to obtain after their completion, and specialized datasets are often not made public, particularly if they contain protected health information (PHI). Studies using datasets from category (i) often reported performances on multiple datasets, consequently making the evaluations more comparable. Overlap of datasets among different studies was relatively low, making comparative analyses difficult. The MIMIC-III (Medical Information Mart for Intensive Care) dataset,[23] was the most frequently used across studies (7/31; 23%), particularly for few-shot classification and NER tasks. This was likely due to the public availability of the dataset and the presence of many labels in it (7000).[24] 6 papers used datasets from shared tasks, of which 4 were from BioNLP,[47,55] one from the Social Media Mining for Health Applications (SMM4H),[41] and one from the Medical Document Anonymization (MEDDOCAN) shared task.[45] Only 3 papers created new datasets, reflecting the paucity of corpora built to support FSL for medical NLP.

**Reconstruction of datasets**

19/31 (61%) reviewed studies reconstructed existing datasets for conducting experiments in few-shot settings (*ie.*, subsets of labeled instances were extracted from larger datasets). For multi-label text classification tasks, especially when the number of labels is very large, and for few-shot NER tasks, reconstructing datasets can be complex. A popular way to represent data in FSL is *K-Shot-N-Way*, where "-shot" applies to the number of examples per category, and the suffix "-way" refers to the number of possible categories. Therefore, *K-Shot-N-Way* means that each of $N$ classes or entities contains $K$ labeled samples, as well as several instances from each class for each test batch. For multi-label classification tasks, each instance may have more than one label, often making it difficult to ensure that the reconstructed datasets include only K labeled samples for each class. Similar challenges exist for NER tasks, as each text segment may have overlapping entities. 39% (12/31) of the studies did not construct special datasets to represent few-shot settings. 16% (5/31) used existing datasets with high class imbalances, and the few-shot algorithms were focused on sparsely-occurring labels.

**Types of methodologies**

23/31 (74%) studies addressed text classification or NER/concept extraction tasks while only 8 (26%) studies focused on others. 24/31 (77%) studies attempted to incorporate *prior knowledge* to augment the small datasets available for training. 19 of these chose to augment the training data with other available annotated datasets as domain knowledge; or through transfer learning, aggregating and adjusting input-output pairs from other larger datasets. For example, due to the scarcity of samples, Manousogiannis et al.[40] attempted to incorporate prior or domain knowledge into their approach by adding concept codes from MEDDRA (Medical Dictionary for Regulatory Activities). 5 papers used pre-trained models learned from other tasks and then refined parameters on the given training data, and 6 studies learned a meta-learner as optimizer or refined meta-learned parameters. it's worth noting that some papers incorporated prior knowledge from more than one source.

**Few-shot text classification**

10/31 studies (33%) focused on few-shot classification, with half of them involving multi-label text classification. Multi-label classification is a popular task because the associated datasets generally contain some very low-frequency classes. 7/10 (70%) papers incorporated data level prior knowledge. 7/10 (70%) classification papers proposed deep learning algorithms, and 3/10 (30%) were inspired by label-wise attention mechanisms. 2/10 (20%) combined few-shot tasks with graphs, such as similarity or co-occurrence graphs, or hierarchical structures that encode relationships between labels for knowledge aggregation. While convolutional neural networks have been popular for FSL, transformer-based models such as BERT[90] and RoBERTa[91] rarely appeared in these articles. Only 1 paper[59] mentioned applying BERT to generate instance embeddings, and then passed top-level output representations into a label-wise attention mechanism.

**Few-shot NER or concept extraction**

8 reviewed papers were described as NER; 5 as concept extraction. Generally, studies described as concept extraction had less commonalities in their methods and involved task-specific configurations based on the characteristics of the data and/or extraction objectives. 5 papers attempted to incorporate data level, 2 model level, and another 2 algorithm level prior knowledge. 63% (5/8) of the studies described as NER employed transfer learning, with training and testing data from different domains. Studies commonly used the BIO (beginning, inside, outside) or IO tagging schemes. 2 papers investigated both BIO and IO tagging schemes, concluding that systems trained using IO schemes outperform those trained using BIO schemes. Studies reported that the O (outside) tag was often ill-defined, as specific entities (*eg.*, time entities such as 'today', 'tomorrow') would be tagged as O if they were not the primary focus of the dataset, while the same entities would be tagged as B or I for other datasets. 5 papers used BIO schemes while 1 considered only the entity names without any tagging schemes. The NLP/machine learning strategies employed varied significantly, and included, for example, the application of fusion layers for combining features,[80] biological semantic and positional features,[84] prototypical representations and nearest neighbor classifiers,[71] transition scorers for modeling transition probabilities between abstract labels,[48,66,71] self-supervised methods,[61,66,83] noise networks for auxiliary training,[54,83] and LSTM cells for encoding multiple entity type sequences.[54]

**Overview of other methods**

6/31 (19%) studies applied meta-learning strategies, and 12/31 (39%) articles demonstrated the advantages of attention mechanisms in few-shot scenarios, such as handling the difficulty of recognizing multiple unseen labels. Among the latter, 5/12 used self-attention-related methods, and 4/12 used label-wise attention mechanisms. 8/31 (26%) studies reproduced prototypical networks, and/or added enhancements to them. Only 1 article used matching networks, and 2 studies included them as baselines.

**Performance metrics**

9/31 (29%) studies used *accuracy*, and the reported values on medical datasets or datasets that included medical texts varied between 67.4% and 96%. Two-thirds (6/9) reported accuracies higher than 70%. For the 17/31 (55%) studies that reported $F_1$-score, performance variations were even larger—from 31.8% to 95.7% (median: 68.6%). We were unable to determine in most cases if the performance differences were due to the effectiveness of the FSL methods, or if the dataset characteristics were primarily responsible.

For the vast majority of studies, reported performances on medical datasets were relatively low compared to other datasets. For papers that reported good performances, we investigated their methods as described, and found that in most cases did not mention how many training examples they used, or their training sizes were large (*e.g.*, in the hundreds). While these approaches may still be considered few-shot learning, comparing these reported performances with those obtained in low-shot settings (*e.g.*, 5-shot) does not constitute fair comparison. We also observed that some of the papers reporting high $F_1$-scores actually included datasets from different domains, and only reported aggregated performances rather than dataset-specific performances.

**DISCUSSION**

In this review, we systematically collected and compared 31 studies that focus on FSL for biomedical NLP. Although there are many potential applications of FSL for biomedical texts, this research space has received relatively little attention. Similar to its progress in the general domain, FSL research in the medical domain has largely been in computer vision.[92] Over two-thirds of the papers included in our review, however, were published in the last 24 months, which illustrates a fast-growing interest. Despite the relatively small number of studies that met our inclusion criteria, several observations were fairly consistent across studies: (i) under the same experimental parameters, the performances reported on medical data were worse than those reported on data from other domains;[35,66,71] (ii) incorporating prior knowledge via transfer learning or using specialized training datasets typically produced better results; and (iii) systems generally reported better performances on datasets with more formal texts compared to those with noisy texts (e.g., from social media).[48,61,71]

We found it difficult to perform head-to-head comparisons of the few proposed methods due to the use of different or non-standardized evaluation strategies, training/test data, and experimental settings. For example, Chalkidis et al.[59] used 50 or less instances in their few-shot setting, while Rios and Kavuluru[21] used 5 or less, making it impossible to perform meaningful comparisons of the proposed methods. In the absence of specialized datasets for FSL, K-Shot-N-Way datasets were commonly reported for simulating few-shot scenarios. In such synthetically created datasets, the number of instances for training are predetermined. Such consistency in characteristics are almost never the case with real-world text-based medical data. Though this design attempts to make direct comparison between different methods or tasks easier, only speculative estimates can be made about how the proposed methods may perform if deployed in real-world settings. It was also typically impossible to compare performances of FSL methods with the state-of-the-art systems reported in prior literature, as FSL methods were expected to underperform compared to methods trained using larger training sets.

Few studies reported the creation of new datasets specialized for FSL, or provided benchmarks that future studies could use for comparison. The scarcity of standardized datasets, and the consequent need to reconstruct datasets for simulating few-shot scenarios is a notable obstacle to progress. Since FSL for biomedical NLP is an underexplored field, such datasets and benchmarks are essential for promoting future development. FSL datasets specialized for biomedical NLP need to contain entities/classes that are naturally sparsely occurring, and the distribution of classes/entities need to reflect real-life data. These conditions are necessary for ensuring that developed systems can be compared directly, and that the system performances reflect what is expected in practical settings. Reconstructed datasets often use randomly sampled subsets for evaluation, making direct comparisons between systems difficult (since the specific training and test instances may not be known), and increasing the potential for biased performance estimates.

**Future directions**

An overarching aim of FSL is to enable systems to learn from few examples, like humans.[93] With text based data, complexities in semantics, syntax, and structure all make it harder to learn and generalize information, particularly with low number of examples. Specialized terminologies used in medical texts present additional challenges for FSL. Our review shows that while the utility of FSL for medical NLP is well acknowledged, this area of research is underexplored. Consequently, FSL systems for biomedical NLP tasks are very much in their infancy, and reported performances are typically low with high variance. Importantly, our review enabled us to identify future research activities that will be most impactful in moving this sub-field of research forward. We outline these in the following subsections.

**Specialized datasets for few-shot learning**
To improve the state of the art in FSL for medical text, the most important activity currently is the creation of specialized, standardized, publicly available datasets. Ideally, such datasets should replicate real-world scenarios and pose practical challenges for FSL. Creation of such datasets will enable the direct comparison of distinct FSL strategies, and of FSL methods with traditional methods (*eg.*, deep neural networks). Public datasets have helped progress NLP and machine learning research over the years, such as through shared tasks.[41] Our review, however, did not find any current shared task that provides specialized datasets for FSL-based biomedical NLP.

**System comparisons and benchmarking**
FSL methods for NLP comprise a wide variety of approaches. Facilitated by standardized datasets, studies need to focus on comparing distinct categories of FSL for biomedical NLP tasks and identify promising methods that need exploration. In the absence of standardized data, benchmarking studies can customize existing datasets and compare distinct FSL methods on identical evaluation sets. Researchers proposing new FSL methods for biomedical NLP should also take the steps necessary to enable head-to-head comparisons and reproducible research. This includes making the evaluation data explicit. Systems evaluated on reconstructed data need to report the exact instances involved in each performance estimate. In the absence of standardized datasets, reporting performances on multiple datasets/settings is also helpful for subjective assessment.

**Opportunities**

The paucity of research in this space means there are many potential opportunities. Domain-independent FSL methods have benefited by incorporating prior knowledge via transfer learning to compensate for the low numbers of training instances.[92] FSL methods for biomedical NLP can follow the same path by using models pre-trained on specific medical datasets. Over the years, medical NLP researchers have created many resources to support NLP methods, such as the Unified Medical Language System (UMLS),[94,95] MedDRA,[96] and others. Effectively incorporating prior knowledge by utilizing these domain-specific knowledge sources is a particularly attractive opportunity.

In the recent review by Wang et al. (2020),[92] the authors specified multi-modal data augmentation as a potential opportunity for improving the state-of-the-art in FSL. The same opportunity also exists in the medical domain. To enable FSL systems achieve performance levels suitable for deployment, future research may focus on augmenting information derived from medical texts with other information, such as images and ontologies. Existing FSL techniques for medical free-text data usually incorporate prior knowledge from one single modality (text), and it is generally not possible to incorporate information from other types of data, such as images. Multi-modal strategies that combine knowledge from distinct sources (*eg.*, texts, images, knowledge bases, ontologies) may enable FSL methods to achieve the performance levels needed to be applicable in real-world medical settings. Intuitively, multi- modal learning models are more akin to human learning. Unsurprisingly data augmentation methods in NLP have recently seen growing interest.[97] Notwithstanding this recent rise, this space is still comparatively underexplored, maybe due to the difficulties in augmentation of natural language data in general, and medical free text in particular, due to the presence of domain-specific terminologies.

There is also the opportunity to create novel datasets specialized for FSL-based biomedical NLP. Creation of comprehensive standardized datasets for FSL involving biomedical texts can lead research in this space. Such contributions to this area at such a crucial point in its progress will inevitably have long-term impact. Efforts to benchmark existing methods on medical datasets will also help identify promising methods requiring further investigation.

## CONCLUSION

FSL approaches have substantial promise for NLP in the medical domain as many medical datasets naturally have low numbers of annotated instances. Some promising approaches have been proposed in the recent past, most of which focused on classification or NER. Meta-learning and transfer learning were commonly used strategies, and a number of studies reported on the benefits of incorporating attention mechanisms. Typical performances of FSL based medical NLP systems are not yet good enough to be suitable for real-world application, and further research on improving performance is required. Lack of public datasets specialized for FSL presents an obstacle to progressing research on the topic, and future research should consider creating such datasets and benchmarks for comparative analyses.

## AUTHOR CONTRIBUTIONS

YGe and AS conducted initial searches and filtering. YGuo, YCY and MAA contributed to the review of the articles, determine their relevance, and/or summarized findings included in the review. All authors contributed to the writing of the final manuscript.

## FUNDING

Research reported in this publication was supported by the National Institute on Drug Abuse (NIDA) of the National Institutes of Health (NIH) under award number R01DA046619. The content is solely the responsibility of the authors and does not necessarily represent the official views of the NIH.

**COMPETING INTERESTS**

None.